\documentclass[10pt, a4paper]{article}
\usepackage{graphicx}
\usepackage{subcaption}
\usepackage{amsmath}
\usepackage[margin=1in]{geometry}
\usepackage{booktabs}
\usepackage{url}
\usepackage{float}
\usepackage{lrec2026} 

\title{Structured Prompting for Arabic Essay Proficiency: A Trait-Centric Evaluation Approach}

\name{
Salim Al Mandhari$^{1}$,
Hieu Pham Dinh$^{2}$,
Mo El-Haj$^{1},^{2}$,
Paul Rayson$^{1},^{2}$
}

\address{
$^{1}$Lancaster University, UK \\
$^{2}$VinUniversity, Vietnam \\
{s.m.almandhari, m.el-haj, p.rayson}@lancaster.ac.uk \\
{24hieu.pd, elhaj.m, paul.r2}@vinuni.edu.vn
}

\abstract{
This paper presents a novel prompt engineering framework for trait specific Automatic Essay Scoring (AES) in Arabic, leveraging large language models (LLMs) under zero-shot and few-shot configurations. Addressing the scarcity of scalable, linguistically informed AES tools for Arabic, we introduce a three-tier prompting strategy (standard, hybrid, and rubric-guided) that guides LLMs in evaluating distinct language proficiency traits such as organization, vocabulary, development, and style. The hybrid approach simulates multi-agent evaluation with trait specialist raters, while the rubric-guided method incorporates scored exemplars to enhance model alignment. 
In zero and few-shot settings, we evaluate eight LLMs on the QAES dataset, the first publicly available Arabic AES resource with trait level annotations. 
Experimental results using Quadratic Weighted Kappa (QWK) and Confidence Intervals show that Fanar-1-9B-Instruct achieves the highest trait level agreement in both zero and few-shot prompting (QWK = 0.28 and CI = 0.41), with rubric-guided prompting yielding consistent gains across all traits and models. Discourse-level traits such as Development and Style showed the greatest improvements. 
These findings confirm that structured prompting, not model scale alone, enables effective AES in Arabic. Our study presents the first comprehensive framework for proficiency oriented Arabic AES and sets the foundation for scalable assessment in low resource educational contexts.
 \\ \newline  }

\begin{document}
\pagestyle{empty}

\maketitleabstract

\section{Introduction}
AES has emerged as a critical tool in educational assessment, enabling scalable and consistent evaluation of language performance. Although substantial progress has been made in AES for high-resource languages such as English, development for Arabic remains limited by the scarcity of annotated datasets, the linguistic complexity of Arabic, and the lack of tools assessing proficiency based on distinct linguistic skills. Existing Arabic AES systems predominantly focus on holistic scoring or content correctness, often neglecting nuanced aspects of proficiency such as organization, vocabulary richness, idea development, and stylistic appropriateness \cite{el-haj-etal-2024-dares,10.1145/3770070}. \citet{ghazawi2024automated}, for example, demonstrated the use of AraBART for Arabic AES, achieving high accuracy in content-based scoring. However, their focus was limited to correctness of responses, with no evaluation of traits or linguistic proficiency. Conversely, English AES research has advanced toward trait scoring and prompt-based evaluation using LLMs \citep{lee2024unleashing, bashendy2024qaes}, yet these innovations have not been fully extended to Arabic due to data and methodological constraints.

To address this gap, our research investigates how prompt engineering techniques including zero-shot and few-shot configurations can be applied to LLMs for trait-specific scoring of Arabic student essays. Rather than relying on fine-tuning or training custom models, we propose a flexible, rubric-informed prompting strategy designed to work directly with existing LLMs\footnote{Code and prompts are available at \url{https://github.com/dinhieufam/Arabic_AES/tree/master}.}

\paragraph{Our contributions are as follows:
}

\begin{itemize}
    \item Our study is the first to evaluate trait-specific prompt engineering strategies on the QAES dataset, using a linguistically aligned hybrid evaluation framework.
    
    \item We introduce a novel three-tier prompting framework for AES in Arabic, composed of: (1) Standard prompts for holistic evaluation, (2) Hybrid trait-based prompts, wherein specialist raters assess specific linguistic features (e.g., organization, vocabulary, tone, development), and (3) Rubric-informed prompts that guide the model using explicit scoring criteria for each trait.

    \item We implement few-shot learning for trait-level prompting, providing example-scored responses to calibrate the model’s judgment. This is, to our knowledge, the first application of few-shot prompting for trait-specific scoring in Arabic.
    \item We evaluate our framework on the QAES dataset, a trait-annotated Arabic AES corpus, and demonstrate competitive alignment with human scores across multiple traits in zero and few-shot settings.
    \item We provide qualitative insights into LLM performance variability across traits and prompt types, revealing which linguistic features are most robustly evaluated by current models.
\end{itemize}

This work demonstrates the feasibility of prompt-based scoring as a scalable and adaptable alternative to traditional supervised AES models, particularly in low-resource settings. It also provides a foundation for the future development of LLM-driven educational tools tailored to the Arabic language.

\section{Related Work}

AES has evolved from rule-based and statistical models to transformer-based language models and advanced prompt engineering. This section categorizes prior work by modeling strategies, dataset development, and prompt-based innovations, emphasizing efforts in Arabic AES.\\

Early AES systems for Arabic utilized string and vector similarity methods. For instance, \citet{abbas2014automated}  developed a Vector Space Model (VSM) to compare 30 essays with reference texts, which was later extended with Latent Semantic Indexing \cite{abbas2015automated}. While foundational, these approaches lacked granularity and linguistic depth. \citet{shehab2018automatic} expanded AES with algorithms such as N-gram and DISCO, achieving moderate accuracy (up to 82\%) on short Arabic responses. \citet{lotfy2023enhanced} further enhanced scoring reliability using ensemble models including Random Forest and Adaboost.\\

Transformer models introduced contextual depth to AES. \citet{ghazawi2024automated} demonstrated this by leveraging AraBART for Arabic AES, achieving 0.88 overall QWK score after training the model across domains using a dataset of over 2,000 essays. Their focus, however, was on evaluating answer correctness in subject-specific contexts rather than linguistic proficiency. Our work, in contrast, centers on assessing language skills through multi-trait analysis.
Sentence-level semantic modeling has been effectively explored by \citet{ramesh2024coherence}, who employed sentence-BERT embeddings with Bi-LSTM and CNNs to capture coherence and cohesion in English AES. The highest QWK score achieved was 0.75. \citet{settles2020machine} combined psychometric modeling (IRT and CAT) with linear ML to adaptively assess language ability in the Duolingo exam framework.\\

Prompt-based AES has gained traction as a fine-tuning-free method. \citet{lee2024unleashing} introduced a zero-shot prompting framework called Multi-Trait Specialization (MTS), enabling LLMs like ChatGPT and LLaMA to achieve competitive trait-level scoring of the TOFEL and ASAP English datasets, gaining QWK scores of 0.43 and 0.35 respectively. The TRATES framework proposed by \citet{eltanbouly2025trates} translated rubrics into guiding questions, improving the scoring alignment of traits using Starling and Gemma models to the average QWK score of 0.59 on the ASAP and ELLIPSE English datasets.
\citet{mansour2024can} tested a four-tier prompt design on English AES using ChatGPT and LLaMA but found these underperformed compared to fine-tuned models, achieving QWK score around 0.57. They did not experiment with few-shot prompting or linguistic trait specialization. \citet{bashendy2024qaes} implemented  feature-based (LR), holistic SOTA (NPCR), single-task learning (STL) and
multi-task learning (MTL) models for both holistic
and trait scoring using AraBERT on the QAES dataset. The highest QWK score was 0.26 achieved by LR, on the holistic scoring,. This finding can be considered as a baseline for our study.\\
Although no work has been found in literature on prompting design for Arabic proficiency scoring, our approach builds on these insights by proposing a three-tier prompting framework: standard, hybrid, and rubric-based for Arabic AES. We incorporate few-shot examples and trait-decomposed scoring, thereby addressing both scoring fidelity and pedagogical interpretability without model fine-tuning.\\

The Arabic AES domain has been limited by resource scarcity. QCAW, introduced by \citet{zaghouani2024qcaw}, provides a bilingual corpus of argumentative essays. QAES, built on QCAW, is the first publicly available Arabic dataset annotated by traits such as organization, vocabulary, and grammar \cite{bashendy2024qaes}. Trait rubrics were derived from official scoring guidelines by the Qatar University Testing Center.
While the only study using QAES relied on logistic regression and AraBERT embeddings \citep{bashendy2024qaes}, it did not explore prompt-based or few-shot approaches. 

\section{Methodology}
\label{sec:methodology}

This study proposes a three-level prompting framework for Arabic AES using large language models (LLMs) in zero-shot and few-shot configurations. Each level is designed to improve linguistic fidelity and alignment with trait-based scoring rubrics, without requiring any model fine-tuning. Table \ref{tab:prompt framework} illustrates the framework structure.\\

\begin{table}[h]
  \centering
  \resizebox{1\columnwidth}{!}{%
  \begin{tabular}{l  p{3cm} p{4cm}}
    \hline
    \textbf{Level} & \textbf{Approach} & \textbf{Key Mechanism} \\
    \hline
    1     & Standard Prompting &	Direct scoring of all traits via a structured zero-shot prompt         \\
    \hline
    2	& Hybrid Trait Prompting	& Multi-agent simulation using domain-specialist raters    \\
        \hline

    3 &	Rubric-Guided Few-Shot Prompting	& Trait-specific prompts with rubric guides and scored examples   \\\hline
  \end{tabular}
  }
   \caption{Overview of Prompting Framework} 
  \label{tab:prompt framework}
\end{table}

The standard prompting level 1 initiates zero-shot prompting by instructing the LLM to assign a score for each of the seven linguistic traits directly based on the student's essay. The prompt template calls for numeric trait scores (e.g., for organization, vocabulary, etc.) and a summed total score. No prior examples or rubrics are provided, and the prompt is general across essays. \\

In the hybrid trait prompting with specialized raters - Level 2, each essay is evaluated by five simulated raters, with each rater specializing in a distinct linguistic trait. Each rater is responsible for scoring only the rubric dimensions that fall within their area of expertise. Table~\ref{tab:rater specializations} presents the five rater specializations. The rubric traits are then mapped to the corresponding rater scores as shown in Table~\ref{tab:rubric mapping}, and the final score is computed by averaging the scores assigned by all relevant raters. A single rater may influence one or more rubric traits, depending on their specialization. This approach ensures that the language model concentrates on a specific linguistic skill during evaluation, thereby simulating the behavior of an expert rater. For instance, the vocabulary specialist focuses exclusively on vocabulary-related aspects and disregards other dimensions such as content development or essay organization.
Equation \ref{rater_map} highlights that the score for each rubric is calculated:

\begin{equation}
    \label{rater_map}
S_j = \frac{1}{|R_j|} \sum_{i \in R_j} s_{ij}
\end{equation}

\noindent Where:
\begin{itemize}
    \item $S_j$: Final score for rubric $j$
    \item $R_j$: Set of raters assigned to rubric $j$
    \item $s_{ij}$: Score given by rater $i$ for rubric $j$
\end{itemize}

This structure promotes focused evaluation while maintaining multi-perspective judgment through overlapping expertise.\\

\begin{table}[h]
  \centering
    \resizebox{0.8\columnwidth}{!}{%
  \begin{tabular}{l  p{2cm} p{4cm}}
    \hline
    \textbf{Rater} & \textbf{Specialization} & \textbf{Evaluation Focus} \\
    \hline
    A	& Organization \& Coherence &	Logical flow, paragraph transitions, structural clarity      \\
            \hline
    B	& Vocabulary \& Lexical Variety &	Word choice, lexical diversity, sophistication, repetition    \\
            \hline
    C	& Grammar, Spelling \& Mechanics &	Punctuation, syntax, spelling, readability \\
            \hline
    D	& Content Development \& Reasoning &	Argument quality, elaboration, evidence use \\
            \hline
    E	& Style, Tone \& Contextual Appropriateness	& Voice, stylistic consistency, audience alignment
    \\\hline
  \end{tabular}
  }
   \caption{The Five Rater Specializations} 
  \label{tab:rater specializations}
\end{table}

\begin{table}[h]
  \centering
  \begin{tabular}{ll}
    \hline
    \textbf{Rubric} & \textbf{Assigned Raters}  \\
    \hline
    Organization	& A, D, C     \\
    Vocabulary	& B, E, C \\
Style	& B, E, C \\
Development	& D, A, B \\
Mechanics	& C \\
Structure	& A, B, C \\
Relevance	& D, B, E
    \\\hline
  \end{tabular}
   \caption{Rubric-to-Rater Mapping} 
  \label{tab:rubric mapping}
\end{table}

Moving to the last prompting level, The rubric-guided few-shot prompting level (level 3) introduces rubric-informed, trait-specific prompts paired with scored examples. For each trait, the model is prompted with a detailed evaluation guide in Arabic, along with a file containing three examples illustrating low, mid, and high scores (e.g., 1, 3, 5). The model compares the target essay to the rubric and examples and produces a trait score with justification in structured JSON format. This method enables greater alignment between model scoring and human evaluation standards, reinforcing clarity, consistency, and explainability across traits. Appendix \ref{prompting templates} highlights the 3 prompting levels designed for the experiments. \\

Although level 1 and level 3 prompts have already been applied in the literature, they are new to Arabic proficiency scoring. The main novelty lies in Prompt 2, which is inspired by human assessment. This reduces the model load and simulates specialist raters. Unlike previous work such as \citet{abbas2014automated}, \citet{abbas2015automated}, and \citet{ghazawi2025well}, where the prompts are unpartitioned (all traits scored in one pass), our method distributes the responsibilities between the simulated raters. The appendix in the camera-ready version will highlight the 3 prompting levels designed for the experiments. 

\section{Experimental Setup}

\subsection{Dataset and Models}
We evaluated our models using the QAES dataset \cite{bashendy2024qaes}. QAES extends the Qatar Corpus of Argumentative Writing (QCAW) by providing fine-grained scores for seven linguistic traits: Organization, Vocabulary, Style, Development, Mechanics, Structure, and Relevance. Trait scores range from 0–5 (except Relevance, which is 0–2), with a total possible score of 32, as shown in Table \ref{tab:qaes-sample}.

\begin{table}[h]
\centering

\small
    \resizebox{\columnwidth}{!}{%
\begin{tabular}{lccccccc|c}
    \hline

\textbf{Essay} & \textbf{Org.} & \textbf{Vocab.} & \textbf{Style} & \textbf{Dev.} & \textbf{Mech.} & \textbf{Struct.} & \textbf{Rel.} & \textbf{Final} \\
    \hline

1 & 4 & 4 & 4 & 4 & 4 & 4 & 2 & 26 \\
2 & 4 & 4 & 4 & 4 & 4 & 4 & 2 & 26 \\
3 & 5 & 4 & 4 & 4 & 4 & 4 & 2 & 27 \\
4 & 5 & 5 & 4 & 4 & 4 & 4 & 2 & 28 \\
    \hline

\end{tabular}
}
\caption{Sample trait annotations for 10 essays in the QAES dataset \cite{bashendy2024qaes}.}
\label{tab:qaes-sample}
\end{table}


Trait annotations were produced by trained human raters following Qatar University’s standardized writing rubrics. Inter-rater agreement was verified through Cohen’s kappa analysis, achieving about 0.72 as an average of the two tasks scored by the two main annotators. \cite{bashendy2024qaes}. We consider this substantial agreement as a threshold for acceptable model performance in our study. The dataset consists of 195 essays written by undergraduate students in response to argumentative prompts in educational settings, primarily focused on technology and communication. Although this dataset is quite small, it was chosen for this project because it is the only dataset available with annotated traits for the assessment of language proficiency in Arabic AES. This resource provides a rich foundation for both zero-shot and few-shot AES modeling in Arabic. 
Other synthetic data by \citet{qwaider2025enhancing} was explored also but found less representative of authentic learner writing. To mitigate the limited dataset, we tested eight LLMs of varying scales across three prompting strategies, focusing on trends consistent across models rather than absolute scores.

Turning to the models, we tested our prompting framework on 8 LLMs based on: (1) model size variety, (2) different foundation bases, and (3) considering efficiency reporting in the literature. We used the corresponding checkpoints for each model available in Hugging Face. Consequently, we selected eight models as follows: ChatGPT-4 \cite{achiam2023gpt}; Fanar-1-9B-Instruct \cite{fanarllm2025}; Jais-family-13B-Chat \cite{sengupta2023jais}; ALLAM-7B-Instruct \cite{bari2024allam}; Qwen1.5-1.8B-Chat \cite{bai2023qwen}; Qwen2.5-7B-Instruct \cite{ahmedqwen}; Qwen3-VL-8B-Instruct \cite{qwen3vl2024}; and LLaMA-2-7B-Chat-hf \cite{touvron2023llama}. These models were selected to test a spectrum of capabilities in open-source and proprietary LLMs with varying levels of Arabic support and instruction tuning.

\subsection{Experimental Protocol}
Our research used the experimental protocol for the three prompting levels described in Section \ref{sec:methodology}. Each essay is evaluated under the same conditions and the predicted scores are compared against the final human annotations. All prompts are in Arabic with rubric-aligned instructions.
Each level is implemented as follows:

\begin{itemize}
    \item Level 1: Single-pass, full-trait zero-shot evaluation.

    \item Level 2: Hybrid multi-agent simulation with trait-specialist raters and rubric-based averaging.

    \item Level 3: Few-shot prompting with trait-specific rubrics and exemplar comparisons.
    
\end{itemize}

Predictions are stored per trait in structured JSON/CSV outputs. All models are evaluated using identical input texts.\\

For the evaluation metric, we used the Quadratic Weighted Kappa (QWK) to measure the agreement between the predicted models and the human gold scores. QWK is defined in Equation \ref{eq:QWK_equation}:
  
\begin{equation}
\label{eq:QWK_equation}
\kappa = 1 - \frac{\sum_{i,j} w_{i,j} O_{i,j}}{\sum_{i,j} w_{i,j} E_{i,j}}, \quad
w_{i,j} = \frac{(i - j)^2}{(N - 1)^2}
\end{equation}

\noindent \textbf{Where:}
\begin{itemize}
    \item $O_{i,j}$: observed agreement matrix
    \item $E_{i,j}$: expected matrix
    \item $w_{i,j}$: penalty for rating difference $|i - j|$
    \item $N$: number of possible score levels for each trait
\end{itemize}

This metric is widely regarded the AES standard and is used to assess the severity of misclassification on an ordinal grading scale. Its use was thoroughly examined by \citet{doewes2023evaluating} and \citet{ghazawi2024automated}, who also discuss its sensitivity to rating scales and best practices for its interpretation.

\section{Results and Analysis}
In this section, we discuss our baseline performance by addressing various scoring dimensions and calculating the QWK score. We compare the 3 prompting levels across the 8 models to demonstrate how our prompt engineering logic effectively enhances the ability of LLMs to assess Arabic language proficiency using zero- and few-shot settings without fine-tuning. In our experiments, both rubric-based and total scores are considered for each essay to evaluate the performance of each model in terms of linguistic skill rubrics and overall assessment quality. Table \ref{tab:qwk_trait_summary} presents the QWK scores across all language traits and the total score for the 8 evaluated models under different prompting levels. It highlights model differences: stronger agreement on Organization, Vocabulary, Development, Mechanics, and Relevance by Fanar and Qwen3-VL. It also shows slight agreement on Style and Structure by the same models. However, smaller models such as Qwen1.5, Qwen2.5, and LLaMA-2, along with models like Jais and ChatGPT-4, exhibit consistently low agreement across all traits. ALLaM performs best among all models in evaluating Structure. Overall, the agreement levels can generally be described as low to fair.

\begin{table*}[ht]
\centering
\small
\begin{tabular}{p{2.5cm}lcccccccc}
\toprule
\textbf{Model} & \textbf{Prompt Level} & \textbf{Org.} & \textbf{Vocab.} & \textbf{Style} & \textbf{Dev.} & \textbf{Mech.} & \textbf{Struct.} & \textbf{Rel.} & \textbf{Total} \\

\midrule
Qwen1.5-1.8B- & L1 & 0.039 & 0.016 & 0.005 & 0.015 & 0.014 & 0.018 & 0.006 & 0.014 \\
                Chat  & L2 & 0.001 & 0.058 & 0.035 & 0.045 & -0.017 & 0.024 & 0.115 & 0.045 \\
                  & L3 & -0.003 & -0.033 & 0.001 & 0.002 & 0.040 & -0.020 & 0.051 & 0.006 \\
\midrule
Qwen2.5-7B- & L1 & 0.015 & 0.016 & 0.074 & 0.015 & 0.067 & 0.007 & 0.00 & 0.021 \\
                  Instruct  & L2 & 0.157 & -0.028 & -0.040 & 0.004 & -0.023 & -0.000 & -0.026 & 0.067 \\
                    & L3 & 0.099 & 0.101 & 0.049 & -0.010 & 0.064 & 0.072 & -0.071 & 0.040 \\
                    \midrule
Qwen3-VL-8B- & L1 & 0.086 & \textbf{0.220} & \textbf{0.191} & 0.045 & 0.188 & 0.096 & 0.000 & 0.125 \\
                  Instruct  & L2 & -0.059 & -0.087 & -0.082 & -0.041 & -0.035 & -0.071 & -0.014 & -0.060 \\
                    & L3 & \textbf{0.264} & 0.001 & 0.179 & 0.061 & 0.194 & 0.176 & 0.206 & 0.206 \\
\midrule
ALLaM-7B- & L1 & 0.143 & -0.037 & 0.051 & -0.001 & -0.031 & 0.015 & 0.021 & 0.039 \\
                        Instruct-preview  & L2 & 0.052 & 0.092 & 0.084 & 0.102 & 0.035 & 0.072 & 0.180 & 0.118 \\
                          & L3 & 0.071 & 0.205 & 0.141 & 0.193 & 0.117 & \textbf{0.198} & 0.141 & 0.264 \\
\midrule
Jais-family-13B- & L1 & -0.006 & 0.002 & 0.000 & 0.026 & 0.010 & -0.006 & 0.077 & 0.007 \\
                    Chat & L2 & 0.006 & 0.004 & 0.006 & 0.009 & 0.006 & 0.003 & 0.003 & 0.008 \\
                     & L3 & 0.000 & 0.000 & 0.000 & 0.000 & 0.000 & 0.000 & 0.000 & 0.000 \\
\midrule
Fanar-1-9B- & L1 & 0.015 & 0.000 & 0.000 & 0.000 & 0.012 & 0.013 & 0.000 & 0.008 \\
                   Instruct & L2 & 0.103 & 0.172 & 0.132 & 0.044 & 0.103 & 0.163 & 0.177 & 0.161 \\
                    & L3 & 0.142 & 0.205 & 0.105 & \textbf{0.204} & \textbf{0.280} & 0.123 & \textbf{0.227} & \textbf{0.284} \\
\midrule
LLaMA-2-7B- & L1 & 0.000 & 0.030 & -0.000 & 0.093 & -0.001 & 0.047 & 0.073 & 0.033 \\
                 Chat-HF  & L2 & 0.015 & 0.039 & 0.044 & 0.091 & -0.000 & 0.034 & 0.030 & 0.046 \\
                     & L3 & 0.000 & 0.000 & 0.000 & 0.000 & 0.000 & 0.000 & 0.000 & 0.000 \\
\midrule
ChatGPT-4 & L1 & 0.090 & 0.043 & 0.012 & 0.055 & 0.060 & 0.035 & 0.000 & 0.050 \\
          & L2 & 0.025 & 0.033 & 0.027 & 0.015 & 0.043 & 0.034 & 0.021 & 0.034 \\
          & L3 & --   & --   & --   & --   & --   & --   & --   & -- \\
\bottomrule
\end{tabular}
\caption{Quadratic Weighted Kappa (QWK) scores for each language trait and total score across all models and prompt levels. GPT-4 results for Prompt Level 3 are omitted due to the high computational cost of running extended evaluations.}

\label{tab:qwk_trait_summary}
\end{table*}

\subsection{Trait and Prompt-Level QWK Scores and Confidence Intervals}
\begin{figure*}[htbp]
    \centering

    \begin{subfigure}[b]{0.46\textwidth}
        \includegraphics[width=\linewidth]{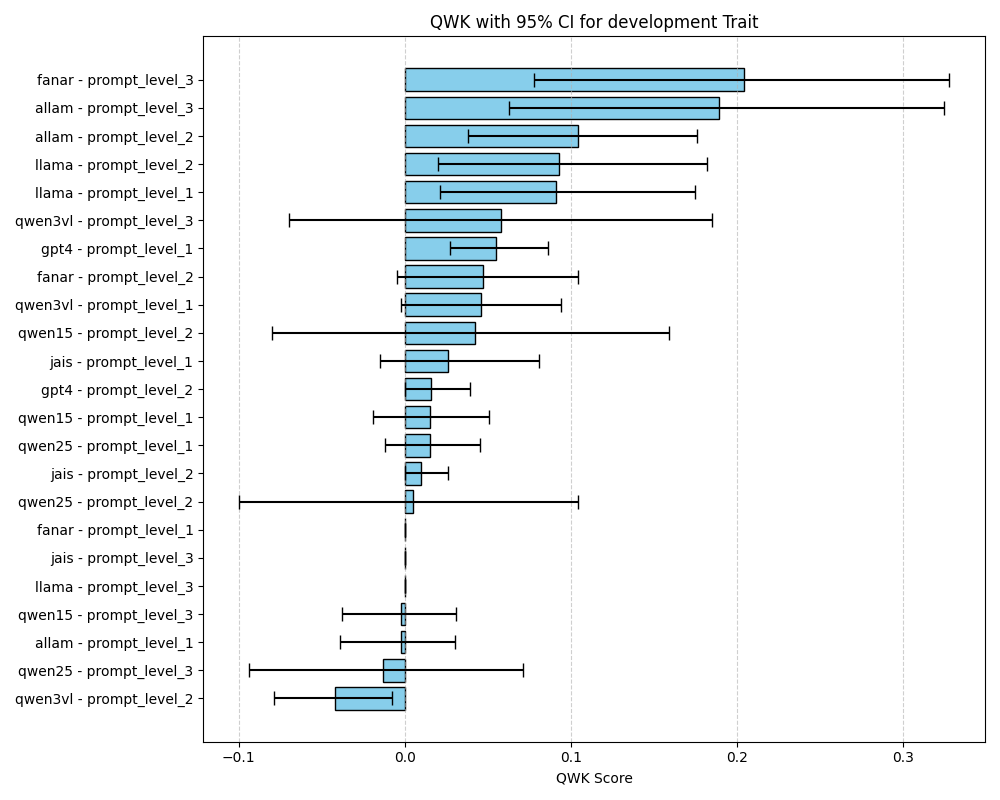}
        \caption{Development}
    \end{subfigure}
    \hfill
    \begin{subfigure}[b]{0.46\textwidth}
        \includegraphics[width=\linewidth]{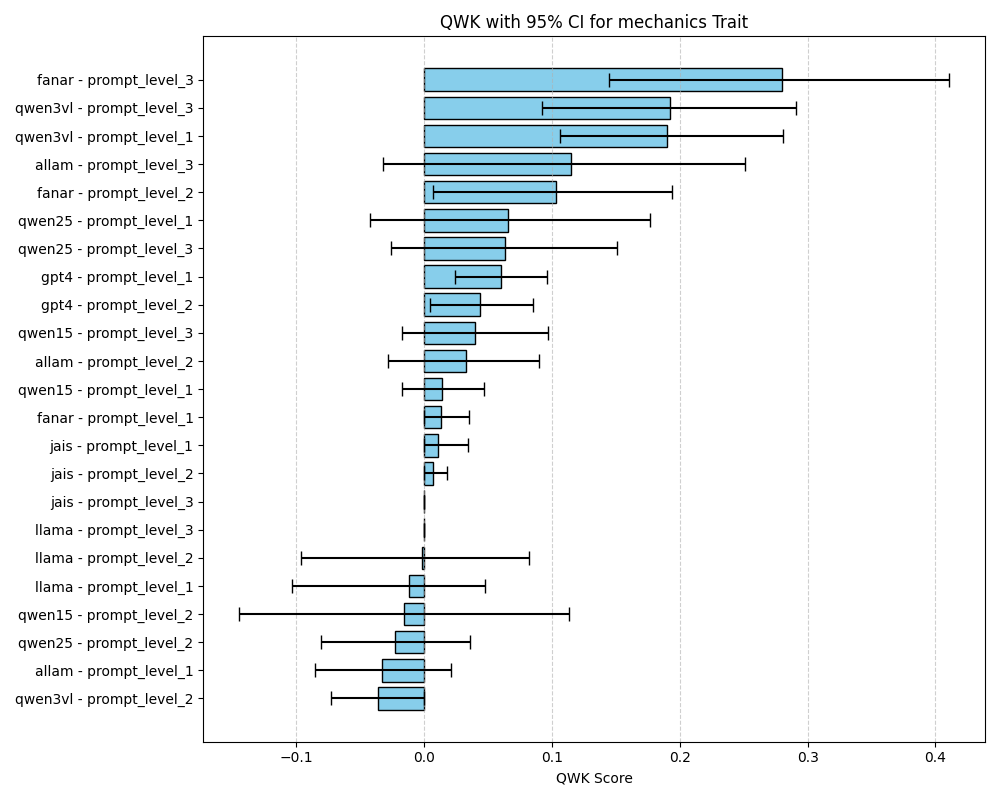}
        \caption{Mechanics}
    \end{subfigure}
    \hfill
    \begin{subfigure}[b]{0.46\textwidth}
        \includegraphics[width=\linewidth]{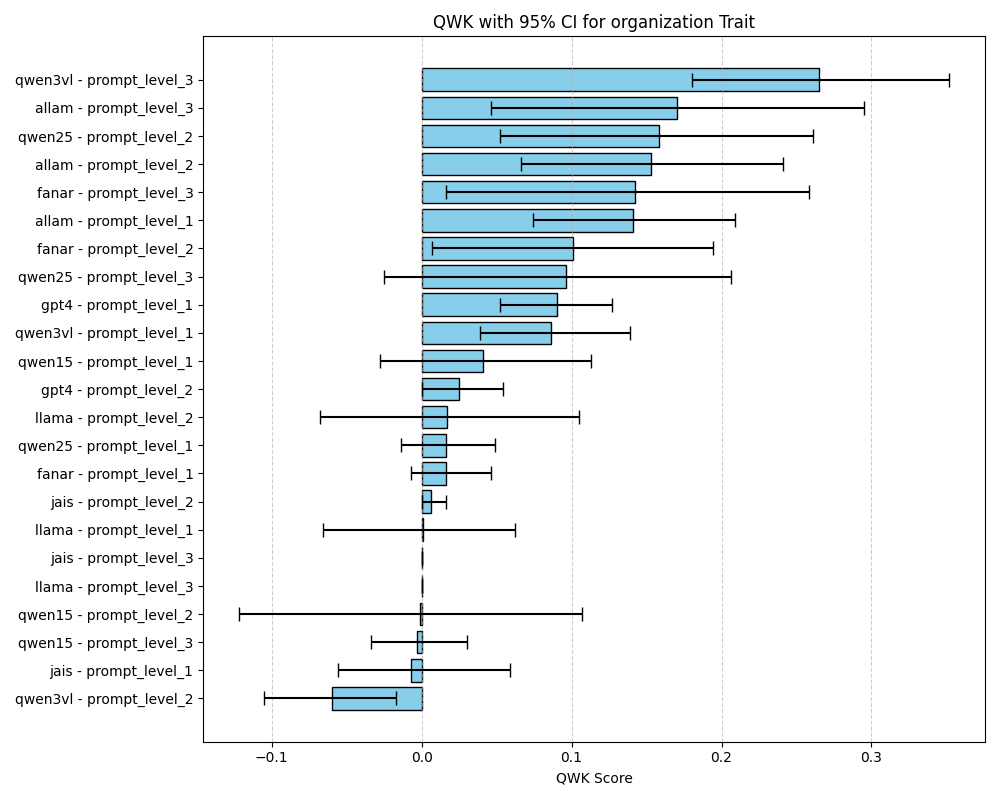}
        \caption{Organization}
    \end{subfigure}
    \hfill
    \begin{subfigure}[b]{0.46\textwidth}
        \includegraphics[width=\linewidth]{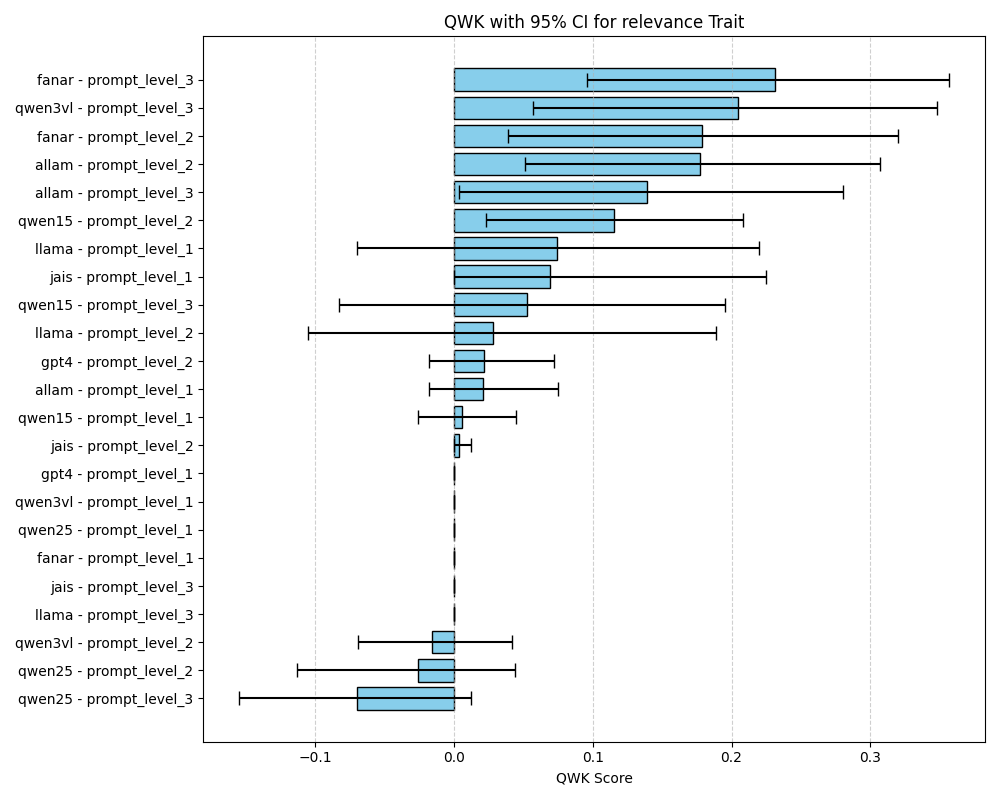}
        \caption{Relevance}
    \end{subfigure}

    \vspace{0.0cm} 

    \begin{subfigure}[b]{0.46\textwidth}
        \includegraphics[width=\linewidth]{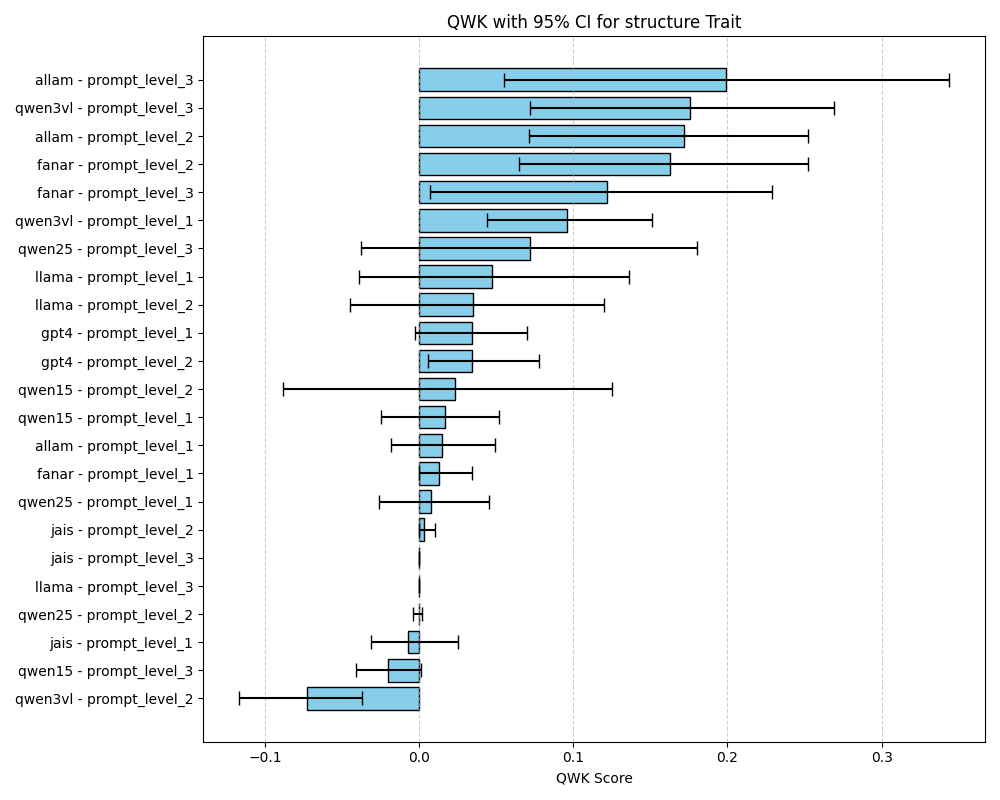}
        \caption{Structure}
    \end{subfigure}
    \hfill
    \begin{subfigure}[b]{0.46\textwidth}
        \includegraphics[width=\linewidth]{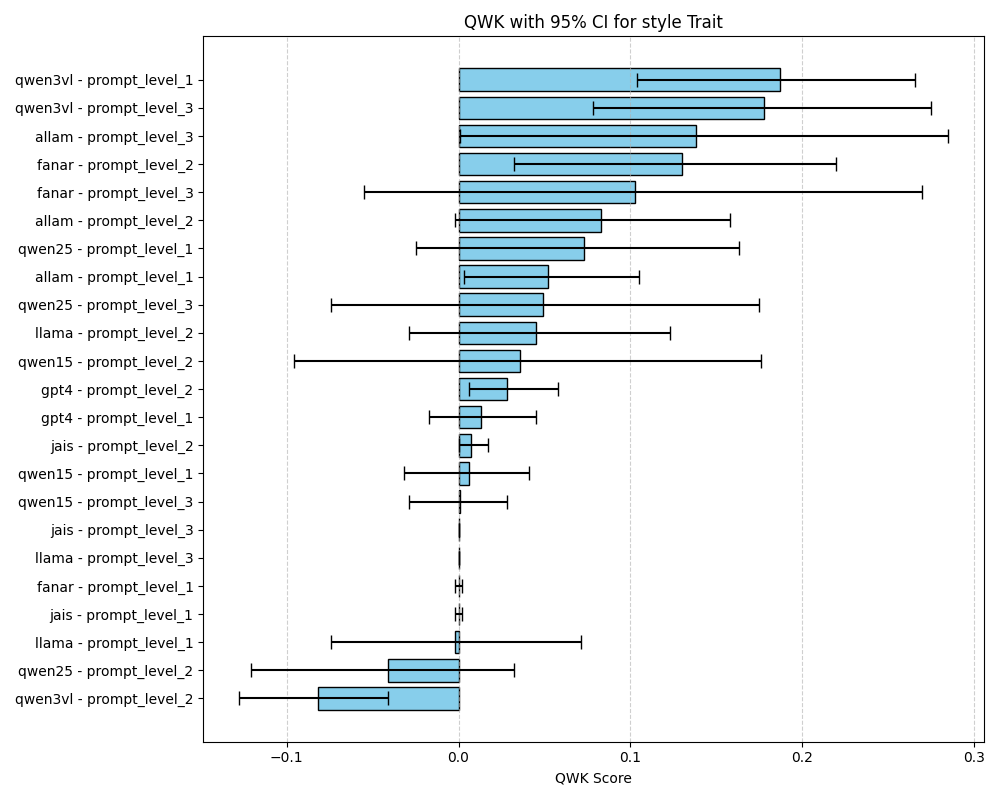}
        \caption{Style}
    \end{subfigure}
    \hfill
    \begin{subfigure}[b]{0.46\textwidth}
        \includegraphics[width=\linewidth]{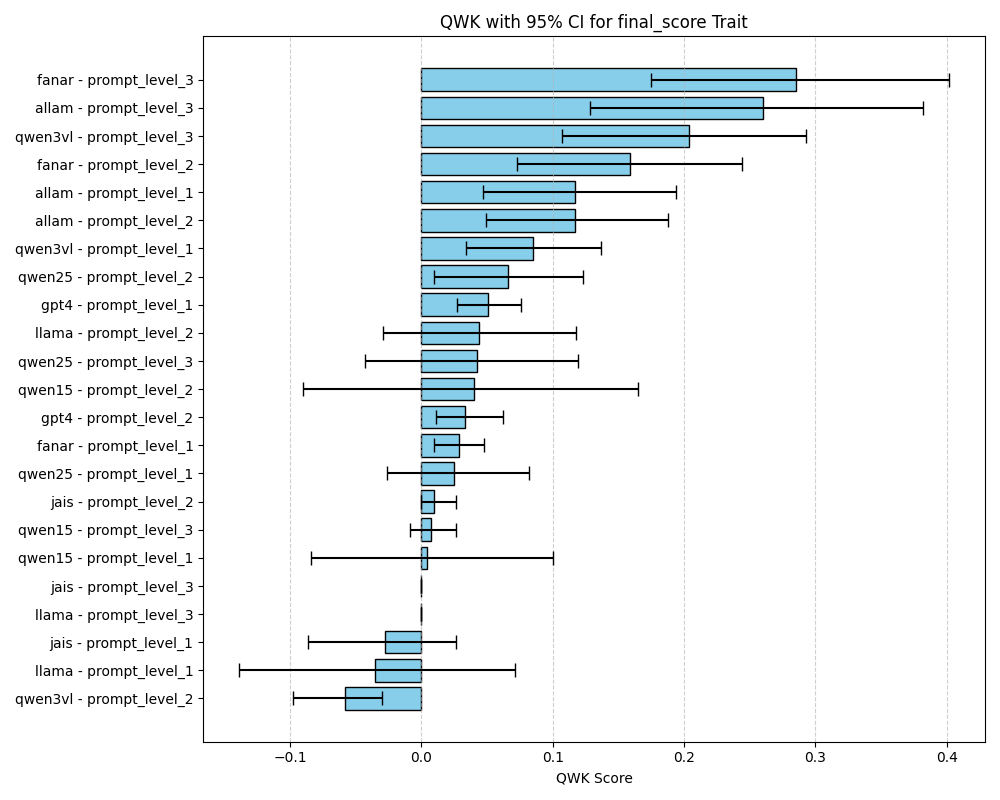}
        \caption{Total}
    \end{subfigure}
    \hfill
    \begin{subfigure}[b]{0.46\textwidth}
        \includegraphics[width=\linewidth]{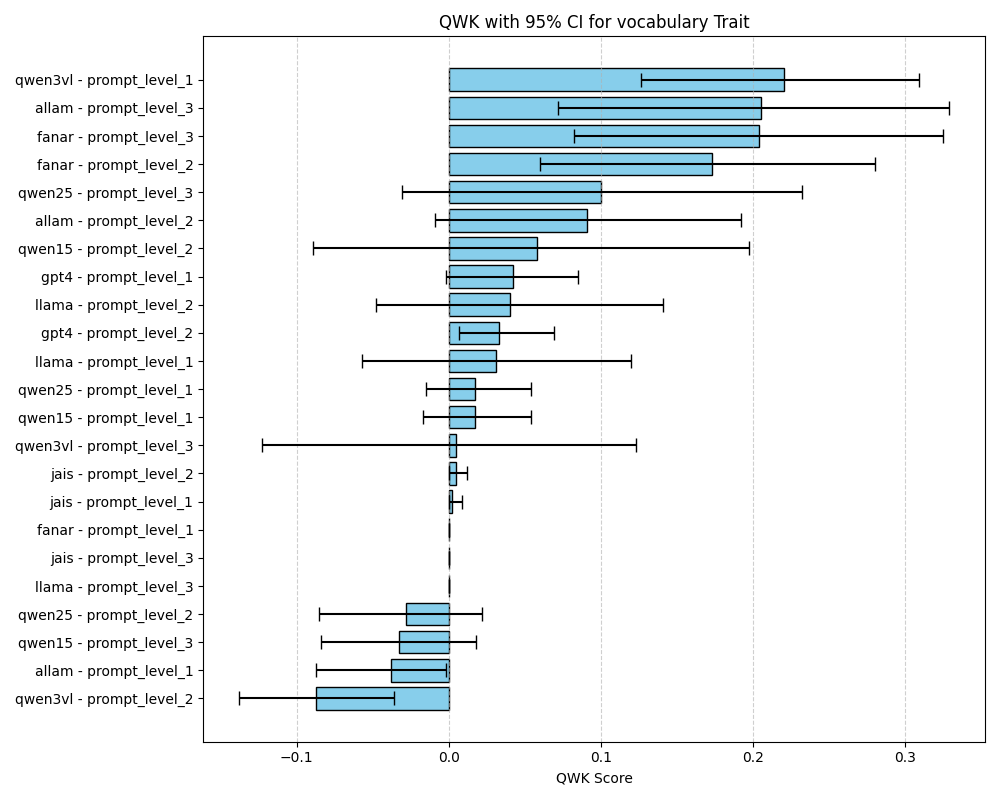}
        \caption{Vocabulary}
    \end{subfigure}

    \caption{QWK Scores with 95\% Confidence Intervals for All Traits}
    \label{fig:qwk_ci_traits}
\end{figure*}

To provide insight beyond a single estimate of QWK scores, we calculate the confidence interval (CI) of all traits cross the models and prompting levels applied. Figure \ref{fig:qwk_ci_traits} presents confidence intervals of each trait, along with its QWK score. \\
The CI offers a statistical range around the observed QWK score that likely contains the true agreement level between the model and human annotations. Specifically, we compute the 95\% CI using non-parametric bootstrapping with 1{,}00 iterations per trait-model-prompt group. This method captures the variability in QWK scores due to the sample of essays evaluated. The CI bounds are calculated as:

\begin{equation}
\text{CI}_{95\%} = \left[ QWK_{2.5}, QWK_{97.5} \right]
\end{equation}
\\
\noindent \textbf{Where:}
 $QWK_{2.5}$ and $QWK_{97.5}$ denote the 2.5th and 97.5th percentiles of the bootstrapped QWK score distribution, respectively \cite{efron1993bootstrap}.

\vspace{0.5em}
\noindent\textbf{Key Observations:}
\begin{itemize}
    \item Performance Trends: Across traits, Fanar and Qwen3-VL consistently achieve the highest QWK scores, primarily under \texttt{prompt\_level\_3}. Vocabulary and Style evaluation in particular is more responsive to \texttt{prompt\_level\_1}. Their CIs are relatively narrow, indicating stable performance.
    \item Prompting Effect: All models show notable improvement from \texttt{prompt\_level\_1} to \texttt{prompt\_level\_3}, confirming that more detailed prompts enhance scoring reliability.
    \item Trait Variability: Traits such as \textit{Vocabulary} and \textit{Relevance} yield higher agreement across models, whereas \textit{Organization}, \textit{Mechanics}, and \textit{Development} exhibit lower QWK scores and wider CIs. Focusing on specific model performance, \textit{Mechanics} appears to be the most responsive trait, with Fanar achieving the highest confidence interval upper bound of \textbf{0.41}.
    \item Low Performing Models: Smaller models such as Qwen1.5, Qwen2.5, and Jais perform inconsistently, often scoring near or below zero on several traits with large CIs, suggesting poor agreement and higher uncertainty. 
\end{itemize}

Comparing our findings to the inter-rater agreement threshold and existing baselines, Table~\ref{tab:threshold_comparison} summarizes the comparative results. The data show that prompt engineering yields a modest improvement in QWK scores, although the results remain considerably below the threshold. Additionally, our study introduces the use of confidence intervals for QWK in Arabic AES, achieving a peak CI upper bound of 0.41, which provides a more robust interpretation of model agreement.\\

\begin{table}[h]
\resizebox{1\columnwidth}{!}{%
\begin{tabular}{p{1.5cm}  p{1.5cm} p{2cm} p{1.5cm}}
\toprule
\textbf{Metric} & \textbf{Baseline} & \textbf{Our Score} & \textbf{Threshold} \\
\midrule
QWK       & 0.26              & 0.28                    & 0.72 \\
CI Upper   & --                & 0.41                    & --   \\
\bottomrule
\end{tabular}}
\caption{Comparison of QWK scores and confidence intervals against the threshold and baseline achieved by QAES.}
\label{tab:threshold_comparison}
\end{table}

\begin{figure}[ht]
    \centering
    \includegraphics[width=0.48\textwidth]{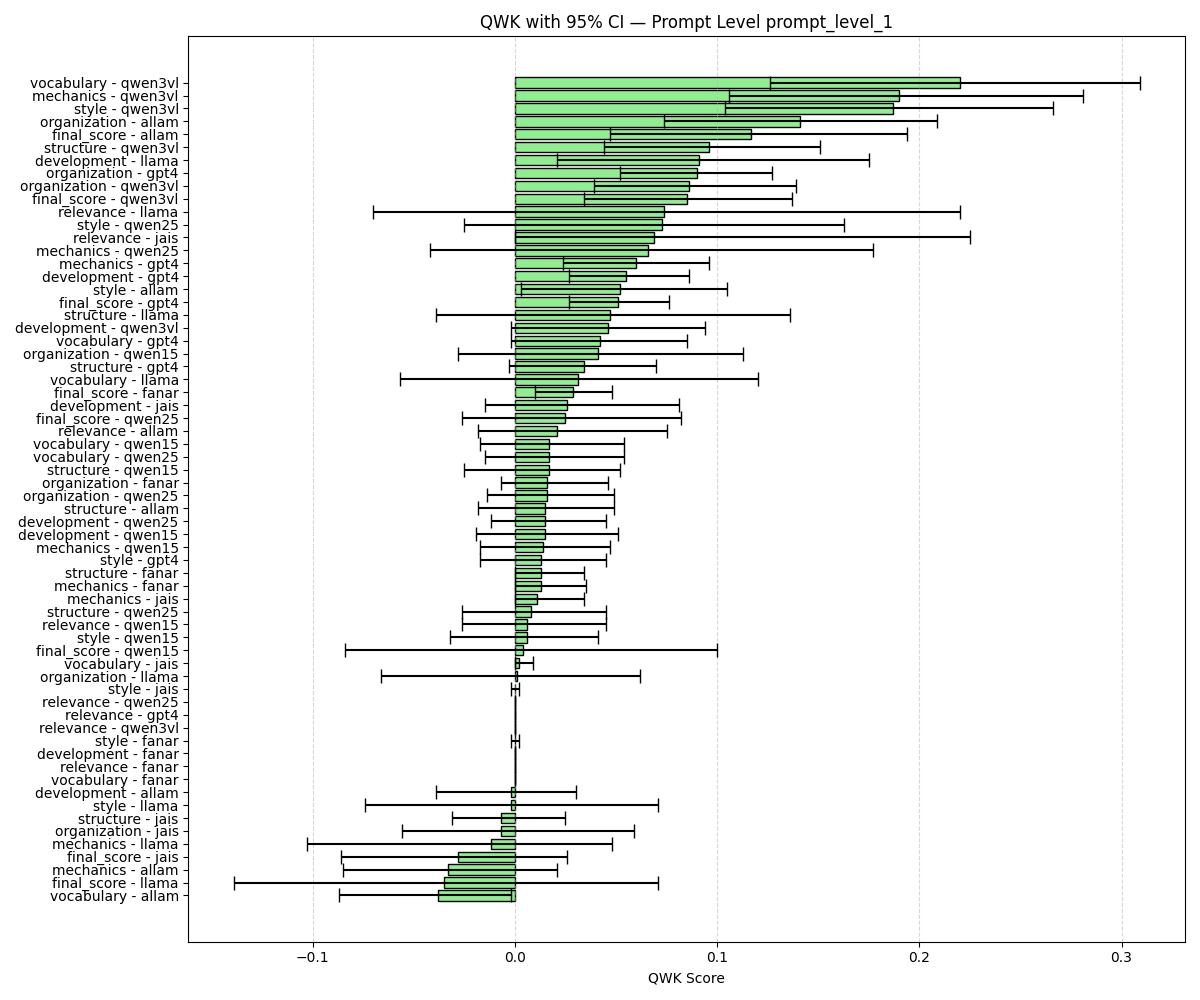}
    \caption{QWK Scores with 95\% Confidence Intervals for Prompt Levels 1}
    \label{fig:qwk_ci_prompt1}
\end{figure}

\begin{figure}[ht]
    \centering
    \includegraphics[width=0.48\textwidth]{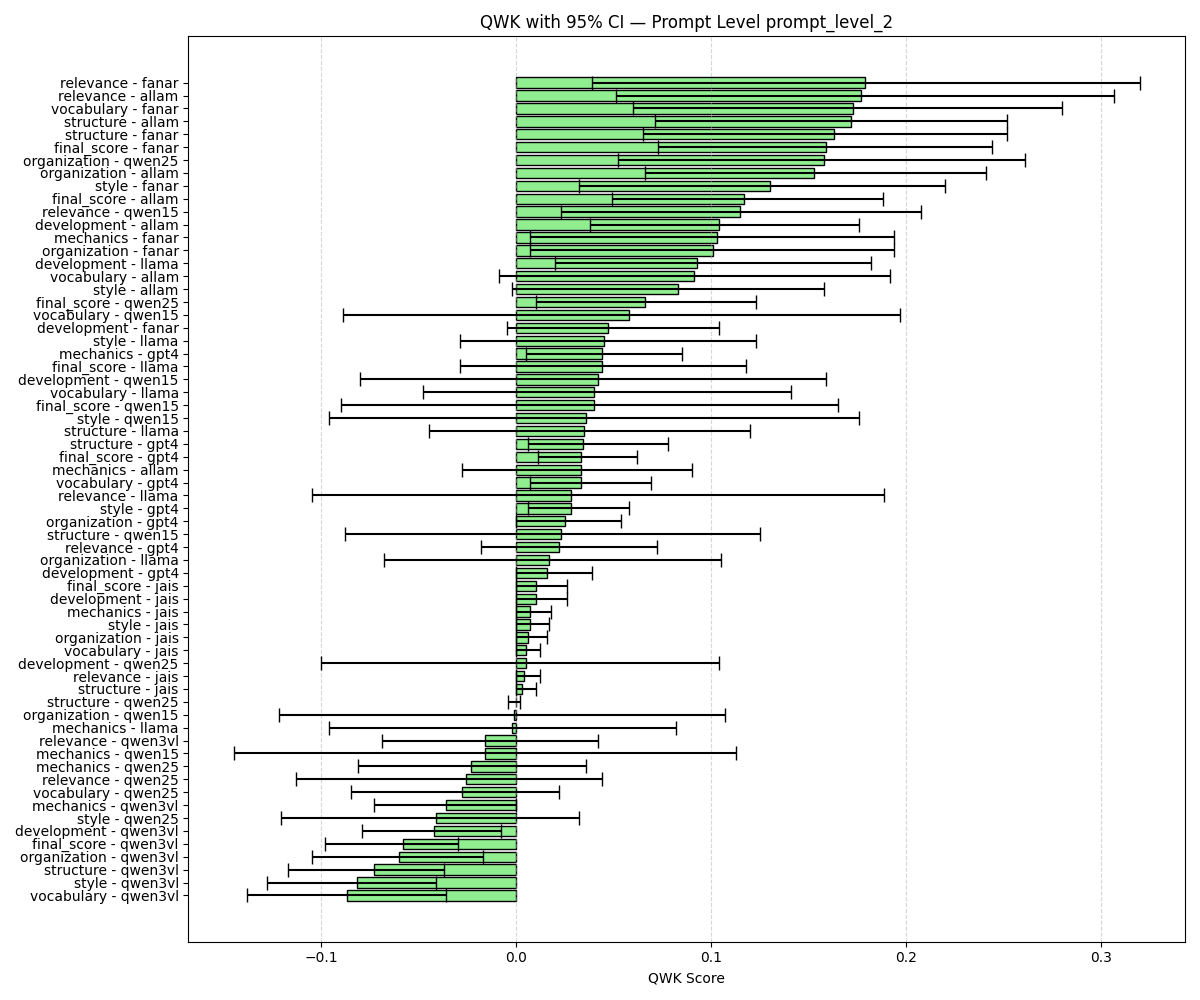}
    \caption{QWK Scores with 95\% Confidence Intervals for Prompt Level 2}
    \label{fig:qwk_ci_prompt2}
\end{figure}

\begin{figure}[ht]
    \centering
    \includegraphics[width=0.48\textwidth]{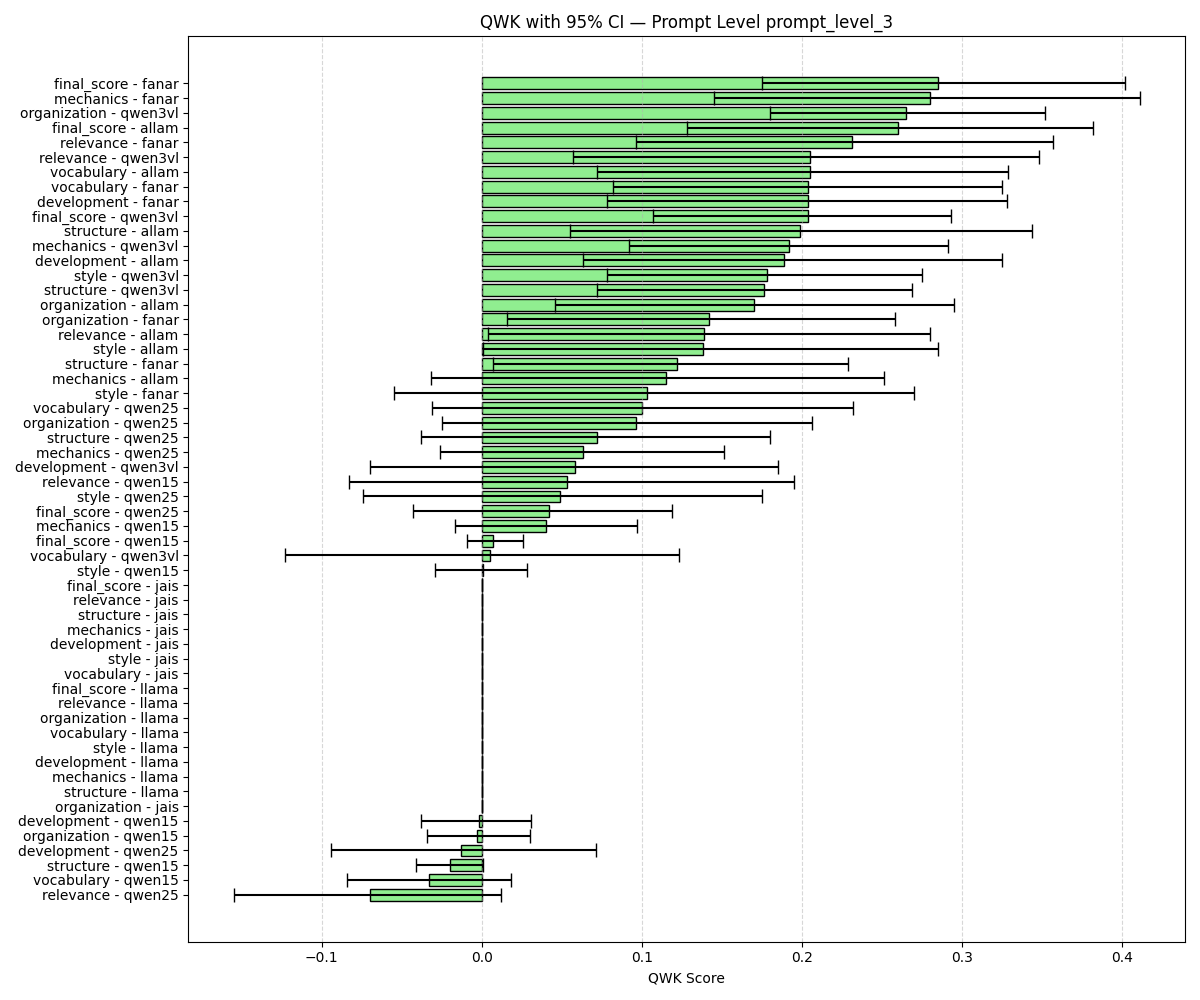}
    \caption{QWK Scores with 95\% Confidence Intervals for Prompt Level 3}
    \label{fig:qwk_ci_prompt3}
\end{figure}

To evaluate the effectiveness of different prompting strategies, we visualize the distribution of QWK scores across all models and prompt levels, along with their corresponding confidence intervals. Figure~\ref{fig:qwk_ci_prompt1}, \ref{fig:qwk_ci_prompt2}, and \ref{fig:qwk_ci_prompt3} present bar plots summarizing QWK scores and 95\% CIs aggregated over the seven linguistic traits and the total score, grouped by prompt level.\\

The results reveal a consistent and interpretable pattern: model performance improves as prompts become more structured and trait-specific. Prompt Level~1, which uses a single holistic instruction, exhibits the widest spread and lowest median scores. In contrast, Prompt Level~3, which leverages rubric-aligned, few-shot trait-specific instructions, achieves the narrowest confidence intervals and the highest QWK scores across most models. These findings underscore that prompting specificity, rather than model size, plays a more critical role in aligning with human-rated scoring.\\

The CI results reveal wide CI across several traits, particularly under Prompt Level 3. This indicates substantial uncertainty in the estimated QWK agreement and suggests high variability in the essay samples used to calculate CI. The wide intervals correspond to larger standard errors, reflecting instability in agreement estimates.
Notably, the Mechanics trait exhibits comparatively narrower CI, suggesting more stable and consistent agreement. This may be attributed to the surface-level nature of mechanical errors, which are less subjective than higher-level discourse traits.
While reducing the confidence level to 90\%, for example, or increasing the sample size would mechanically narrow the interval width, this would not resolve the underlying variability in agreement.

\subsection{Interpretation and Discussion}

These results demonstrate that prompt structure has a greater impact on AES performance than model size alone, particularly for complex linguistic traits. The most significant QWK improvements were observed with the hybrid prompting method operating in a zero-shot setting, as reflected in the following:

\begin{itemize}

    \item Most of the traits, Organization, Vocabulary, Style, and Relevance, as well as the total score, showed improvement in most LLM runs. In contrast, other traits such as Development, Mechanics, and Structure exhibited only moderate improvement.
    
    \item The Fanar-1-9B and ALLaM-7B models demonstrated the strongest instruction tuning and the most robust Arabic language coverage, with consistently better understanding and scoring in all traits.

\end{itemize}
When examining the effect of few-shot settings within Trait Prompting, the Structure trait was the only one to show notable improvement. In contrast, the impact on the other traits and the overall score was minimal.

Another notable observation is that models showing a performance drop across prompt levels also tend to have very low QWK scores. This supports the hypothesis that as prompt complexity increases, smaller models struggle more with comprehension. Interestingly, for these smaller models, scores for Prompt 2 are often higher than for Prompt 1. This could be because Prompt 1 requires 8 trait scores, while Prompt 2 requires only 5 (A, B, C, D, and E).

Additionally, for Prompt 3, both LLaMA and JAIS have a QWK score of 0, as they failed to generate meaningful outputs from which scores could be extracted. This further illustrates the difficulty these models face with more complex prompts.

Assessing language proficiency is considerably more complex than evaluating content correctness in AES. \citet{ghazawi2025well} reported a QWK of 0.45 for GPT-4 in zero-shot content scoring, while in our study, the same model achieved only 0.05 in Prompt Level 1 for trait-based proficiency scoring. This highlights the added difficulty of evaluating traits such as organization and development, which require deeper contextual and rubric-aligned understanding.

These findings support our hypothesis that prompt engineering, especially rubric-aligned and example-based prompting, can transform general-purpose LLMs into effective AES tools for Arabic, without any fine-tuning or domain adaptation.

\section{Conclusion and Future Work}
This paper proposed a prompt-based framework for trait-specific Arabic AES using large language models. Through a three-level prompting strategy, standard, hybrid, and rubric-guided, we showed that carefully designed prompts can elicit reliable trait-level scoring in zero and few-shot settings. Experiments on eight LLMs using the QAES dataset confirmed that prompt engineering improves performance, especially for small and medium models suitable for educational use where resources and privacy matter. Our results also reveal that traits such as development, organisation, and style remain challenging, requiring structured linguistic guidance rather than simple holistic scoring.

For future work, we plan to expand the dataset to improve robustness and generalisability and explore the transferability of our framework to non-Arabic AES tasks. We also aim to address methodological challenges such as rater disagreement and score aggregation in hybrid prompting. This work provides a foundation for scalable, low-resource AES solutions in Arabic and other underrepresented languages.

\section*{Limitations}
Although our framework demonstrates strong performance in trait-specific Arabic AES, several limitations remain. The scarcity of annotated Arabic AES datasets restricts large-scale evaluation and limits linguistic and demographic coverage. In addition, reliance on proprietary models like GPT-4 incurs high cost, making large-scale deployment difficult in resource-constrained environments.

These limitations highlight the need for more accessible Arabic AES datasets and optimisation strategies for cost-effective implementation. Future work should also explore open-source alternatives and parameter-efficient adaptation techniques to improve practicality and reproducibility.

\nocite{*}
\section{Bibliographical References}\label{sec:reference}

\bibliographystyle{lrec2026-natbib}
\bibliography{lrec2026-example}

@misc{fanarllm2025,
      title={Fanar: An Arabic-Centric Multimodal Generative AI Platform}, 
      author={Fanar Team and Ummar Abbas and Mohammad Shahmeer Ahmad and Firoj Alam and Enes Altinisik and Ehsannedin Asgari and Yazan Boshmaf and Sabri Boughorbel and Sanjay Chawla and Shammur Chowdhury and Fahim Dalvi and Kareem Darwish and Nadir Durrani and Mohamed Elfeky and Ahmed Elmagarmid and Mohamed Eltabakh and Masoomali Fatehkia and Anastasios Fragkopoulos and Maram Hasanain and Majd Hawasly and Mus'ab Husaini and Soon-Gyo Jung and Ji Kim Lucas and Walid Magdy and Safa Messaoud and Abubakr Mohamed and Tasnim Mohiuddin and Basel Mousi and Hamdy Mubarak and Ahmad Musleh and Zan Naeem and Mourad Ouzzani and Dorde Popovic and Amin Sadeghi and Husrev Taha Sencar and Mohammed Shinoy and Omar Sinan and Yifan Zhang and Ahmed Ali and Yassine El Kheir and Xiaosong Ma and Chaoyi Ruan},
      year={2025},
      url={https://arxiv.org/abs/2501.13944}, 
}

@article{achiam2023gpt,
  title={Gpt-4 technical report},
  author={Achiam, Josh and Adler, Steven and Agarwal, Sandhini and Ahmad, Lama and Akkaya, Ilge and Aleman, Florencia Leoni and Almeida, Diogo and Altenschmidt, Janko and Altman, Sam and Anadkat, Shyamal and others},
  journal={arXiv preprint arXiv:2303.08774},
  year={2023}
}

@article{sengupta2023jais,
  title={Jais and jais-chat: Arabic-centric foundation and instruction-tuned open generative large language models},
  author={Sengupta, Neha and Sahu, Sunil Kumar and Jia, Bokang and Katipomu, Satheesh and Li, Haonan and Koto, Fajri and Marshall, William and Gosal, Gurpreet and Liu, Cynthia and Chen, Zhiming and others},
  journal={arXiv preprint arXiv:2308.16149},
  year={2023}
}

@article{10.1145/3770070,
author = {Almujaiwel, Sultan and Premasiri, Damith and Ranasinghe, Tharindu and El-Haj, Mo and Mitkov, Ruslan},
title = {Complex Concept-Based Readability Estimation from Arabic Curriculum},
year = {2025},
publisher = {Association for Computing Machinery},
address = {New York, NY, USA},
issn = {2375-4699},
url = {https://doi.org/10.1145/3770070},
doi = {10.1145/3770070},
note = {Just Accepted},
journal = {ACM Trans. Asian Low-Resour. Lang. Inf. Process.},
month = oct
}

@inproceedings{el-haj-etal-2024-dares,
    title = "{DARES}: Dataset for {A}rabic Readability Estimation of School Materials",
    author = "El-Haj, Mo  and
      Almujaiwel, Sultan  and
      Premasiri, Damith  and
      Ranasinghe, Tharindu  and
      Mitkov, Ruslan",
    editor = "Nunzio, Giorgio Maria Di  and
      Vezzani, Federica  and
      Ermakova, Liana  and
      Azarbonyad, Hosein  and
      Kamps, Jaap",
    booktitle = "Proceedings of the Workshop on DeTermIt! Evaluating Text Difficulty in a Multilingual Context @ LREC-COLING 2024",
    month = may,
    year = "2024",
    address = "Torino, Italia",
    publisher = "ELRA and ICCL",
    url = "https://aclanthology.org/2024.determit-1.10/",
    pages = "103--113"
}

@article{bari2024allam,
  title={Allam: Large language models for arabic and english},
  author={Bari, M Saiful and Alnumay, Yazeed and Alzahrani, Norah A and Alotaibi, Nouf M and Alyahya, Hisham A and AlRashed, Sultan and Mirza, Faisal A and Alsubaie, Shaykhah Z and Alahmed, Hassan A and Alabduljabbar, Ghadah and others},
  journal={arXiv preprint arXiv:2407.15390},
  year={2024}
}

@article{bai2023qwen,
  title={Qwen technical report},
  author={Bai, Jinze and Bai, Shuai and Chu, Yunfei and Cui, Zeyu and Dang, Kai and Deng, Xiaodong and Fan, Yang and Ge, Wenbin and Han, Yu and Huang, Fei and others},
  journal={arXiv preprint arXiv:2309.16609},
  year={2023}
}

@article{ahmedqwen,
  title={Qwen 2.5: A Comprehensive Review of the Leading Resource-Efficient LLM with potentioal to Surpass All Competitors},
  author={Ahmed, Imtiaz and Islam, Sadman and Datta, Partha Protim and Kabir, Imran and Chowdhury, Naseef Ur Rahman and Haque, Ahshanul},
  journal={Authorea Preprints},
  publisher={Authorea},
  year={2025}

}

@article{touvron2023llama,
  title={Llama 2: Open foundation and fine-tuned chat models},
  author={Touvron, Hugo and Martin, Louis and Stone, Kevin and Albert, Peter and Almahairi, Amjad and Babaei, Yasmine and Bashlykov, Nikolay and Batra, Soumya and Bhargava, Prajjwal and Bhosale, Shruti and others},
  journal={arXiv preprint arXiv:2307.09288},
  year={2023}
}

@article{ghazawi2024automated,
  title={Automated essay scoring in Arabic: a dataset and analysis of a BERT-based system},
  author={Ghazawi, Rayed and Simpson, Edwin},
  journal={arXiv preprint arXiv:2407.11212},
  year={2024}
}

@inproceedings{bashendy2024qaes,
  title={QAES: First Publicly-Available Trait-Specific Annotations for Automated Scoring of Arabic Essays},
  author={Bashendy, May and Albatarni, Salam and Eltanbouly, Sohaila and Zahran, Eman and Elhuseyin, Hamdo and Elsayed, Tamer and Massoud, Walid and Bouamor, Houda},
  booktitle={Proceedings of The Second Arabic Natural Language Processing Conference},
  pages={337--351},
  year={2024}
}

@inproceedings{doewes2023evaluating,
  title={Evaluating quadratic weighted kappa as the standard performance metric for automated essay scoring},
  author={Doewes, Afrizal and Kurdhi, Nughthoh and Saxena, Akrati},
  booktitle={16th International Conference on Educational Data Mining, EDM 2023},
  pages={103--113},
  year={2023},
  organization={International Educational Data Mining Society (IEDMS)}
}

@article{ramesh2024coherence,
  title={Coherence-based automatic short answer scoring using sentence embedding},
  author={Ramesh, Dadi and Sanampudi, Suresh Kumar},
  journal={European Journal of Education},
  volume={59},
  number={3},
  pages={e12684},
  year={2024},
  publisher={Wiley Online Library}
}

@article{settles2020machine,
  title={Machine learning--driven language assessment},
  author={Settles, Burr and T. LaFlair, Geoffrey and Hagiwara, Masato},
  journal={Transactions of the Association for computational Linguistics},
  volume={8},
  pages={247--263},
  year={2020},
  publisher={MIT Press One Rogers Street, Cambridge, MA 02142-1209, USA journals-info~…}
}

@inproceedings{zaghouani2024qcaw,
  title={Qcaw 1.0: building a qatari corpus of student argumentative writing},
  author={Zaghouani, Wajdi and Ahmed, Abdelhamid and Zhang, Xiao and Rezk, Lameya},
  booktitle={Proceedings of the 2024 Joint International Conference on Computational Linguistics, Language Resources and Evaluation (LREC-COLING 2024)},
  pages={13382--13394},
  year={2024}
}

@article{lee2024unleashing,
  title={Unleashing Large Language Models' Proficiency in Zero-shot Essay Scoring},
  author={Lee, Sanwoo and Cai, Yida and Meng, Desong and Wang, Ziyang and Wu, Yunfang},
  journal={arXiv preprint arXiv:2404.04941},
  year={2024}
}

@article{abbas2014automated,
  title={Automated Arabic essay scoring (AAES) using vector space model (VSM)},
  author={Abbas, Ayad R and Al-qaza, Ahmed S},
  journal={Journal of Al-Turath University College},
  volume={15},
  pages={25--39},
  year={2014}
}

@article{abbas2015automated,
  title={Automated arabic essay scoring (aaes) using vectors space model (vsm) and latent semantics indexing (lsi)},
  author={Abbas, Ayad R and Al-qazaz, Ahmed S},
  journal={Eng. Technol. J.},
  volume={33},
  number={3},
  pages={410--426},
  year={2015}
}

@article{shehab2018automatic,
  title={An automatic Arabic essay grading system based on text similarity Algorithms},
  author={Shehab, Abdulaziz and Faroun, Mahmoud and Rashad, Magdi},
  journal={International Journal of Advanced Computer Science and Applications},
  volume={9},
  number={3},
  year={2018},
  publisher={Science and Information (SAI) Organization Limited}
}

@article{lotfy2023enhanced,
  title={An Enhanced Automatic Arabic Essay Scoring System Based on Machine Learning Algorithms},
  author={Lotfy, Nourmeen and Shehab, Abdulaziz and Elhoseny, Mohammed and Abu-Elfetouh, Ahmed},
  journal={CMC-COMPUTERS MATERIALS \& CONTINUA},
  volume={77},
  number={1},
  pages={1227--1249},
  year={2023},
  publisher={TECH SCIENCE PRESS 871 CORONADO CENTER DR, SUTE 200, HENDERSON, NV 89052 USA}
}

@article{eltanbouly2025trates,
  title={TRATES: Trait-Specific Rubric-Assisted Cross-Prompt Essay Scoring},
  author={Eltanbouly, Sohaila and Albatarni, Salam and Elsayed, Tamer},
  journal={arXiv preprint arXiv:2505.14577},
  year={2025}
}

@article{mansour2024can,
  title={Can large language models automatically score proficiency of written essays?},
  author={Mansour, Watheq and Albatarni, Salam and Eltanbouly, Sohaila and Elsayed, Tamer},
  journal={In Proceedings of the 2024 Joint International Conference on Computational Linguistics, Language Resources and Evaluation (LREC-COLING 2024), pages 2777–2786, Torino, Italia. ELRA and ICCL.},
  year={2024}
}

@article{ghazawi2025well,
  title={How well can LLMs Grade Essays in Arabic?},
  author={Ghazawi, Rayed and Simpson, Edwin},
  journal={arXiv preprint arXiv:2501.16516},
  year={2025}
}

@book{efron1993bootstrap,
  title={An Introduction to the Bootstrap},
  author={Efron, Bradley and Tibshirani, Robert J.},
  year={1993},
  publisher={Chapman \& Hall/CRC}
}

@article{qwaider2025enhancing,
  title={Enhancing arabic automated essay scoring with synthetic data and error injection},
  author={Qwaider, Chatrine and Alhafni, Bashar and Chirkunov, Kirill and Habash, Nizar and Briscoe, Ted},
  journal={arXiv preprint arXiv:2503.17739},
  year={2025}
}

@misc{qwen3vl2024,
  author       = {QwenLM Team},
  title        = {{Qwen3-VL: Multimodal Large Language Model}},
  year         = {2024},
  howpublished = {\url{https://github.com/QwenLM/Qwen3-VL}},
  note         = {Accessed: October 20, 2025}
}

\label{lr:ref}

\section*{Appendix}
\appendix

\section{Prompting Templates}
\label{prompting templates}

The following templates represent the three common prompting designs used to run all the LLMs.

\begin{figure*}[t]
  \includegraphics[width=1.1\linewidth]{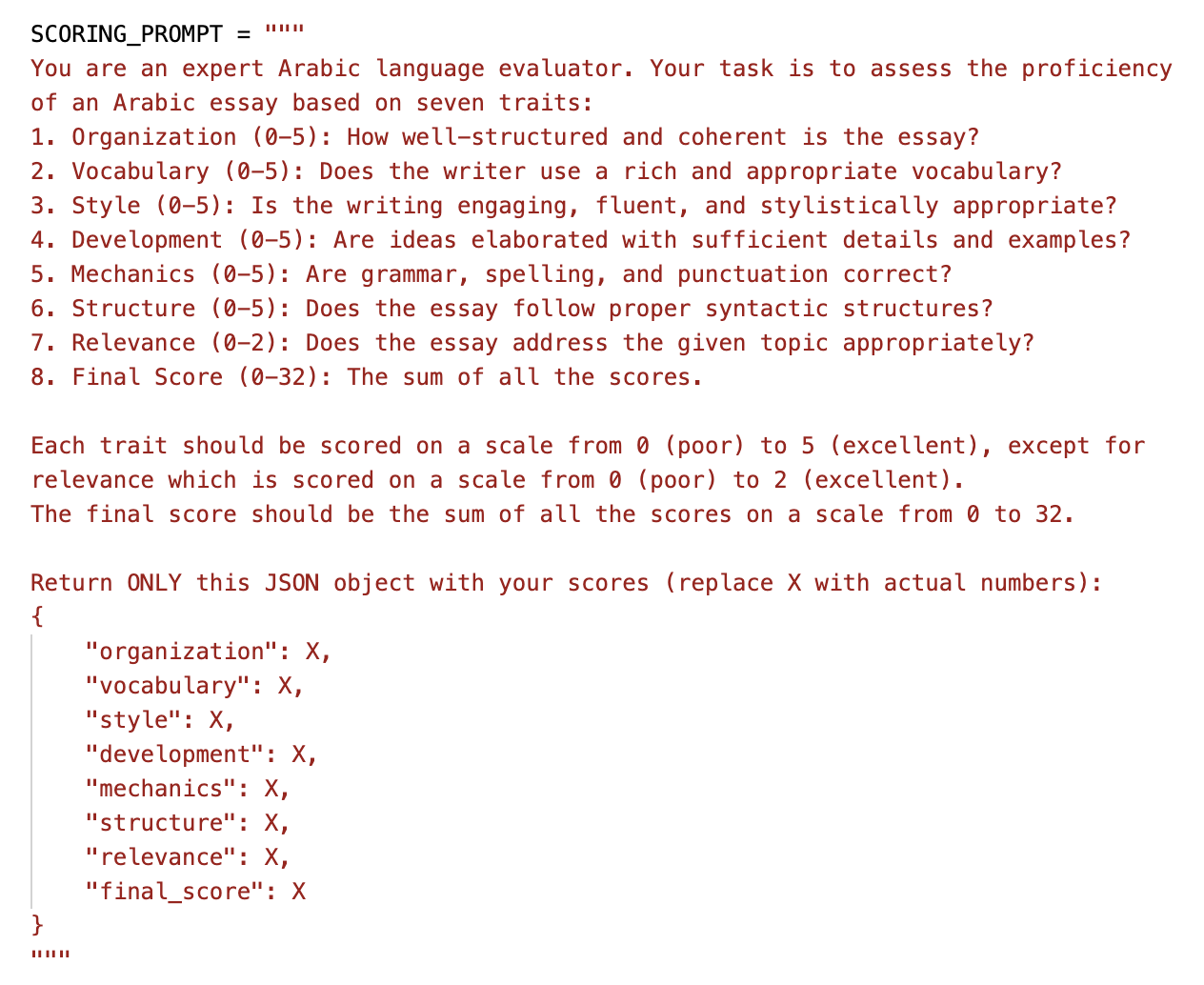} \hfill

\caption{Standard Prompt - Level 1}

\end{figure*}

\begin{figure*}[t]

\includegraphics[width=1\linewidth]{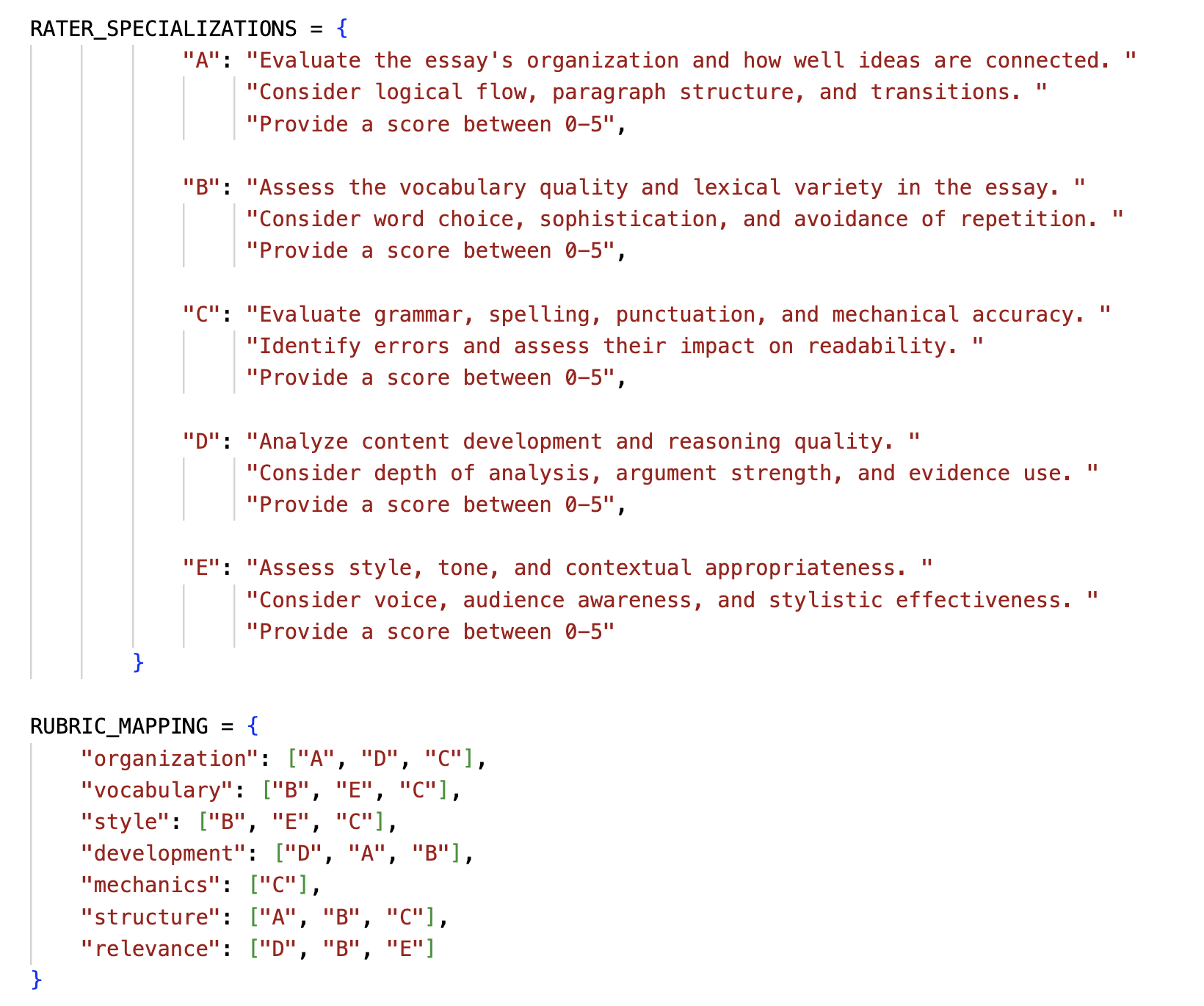}
\caption{Hybrid Trait Prompt - Level 2 (first part)}

\end{figure*}

\begin{figure*}[t]

\includegraphics[width=1\linewidth]{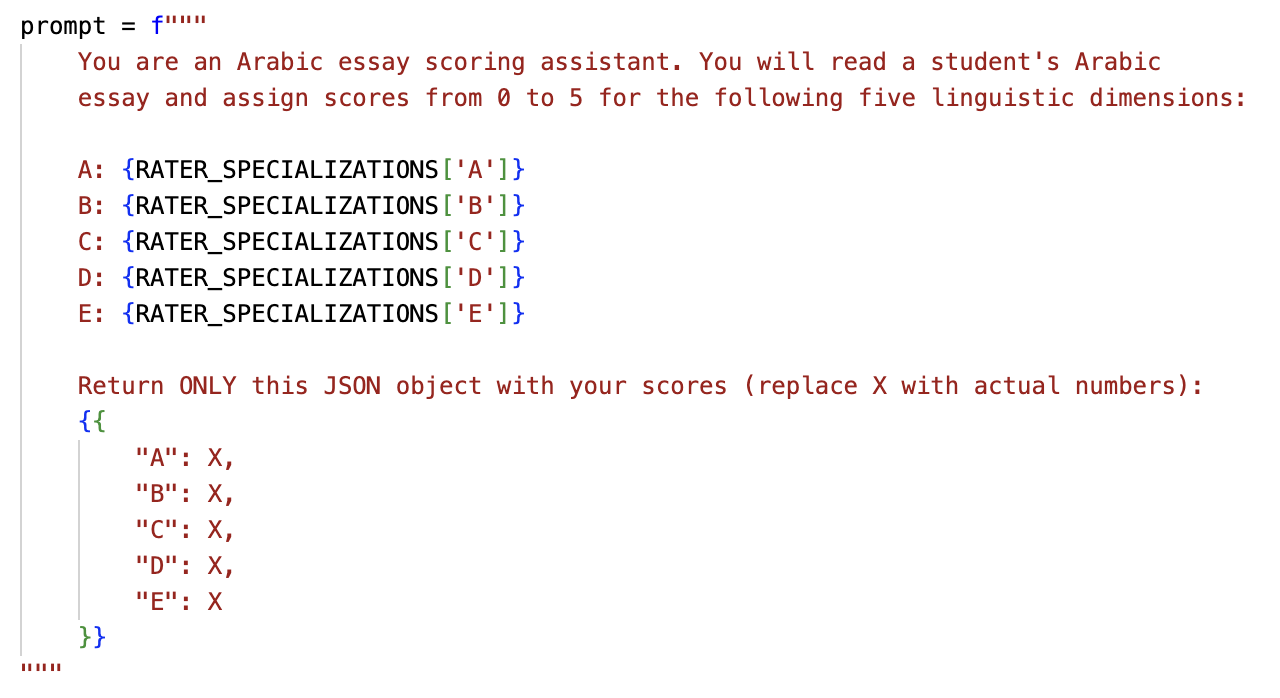}
\caption{Hybrid Trait Prompt - Level 2 (second part)}

\end{figure*}

\begin{figure*}[t]\includegraphics[width=1\linewidth]{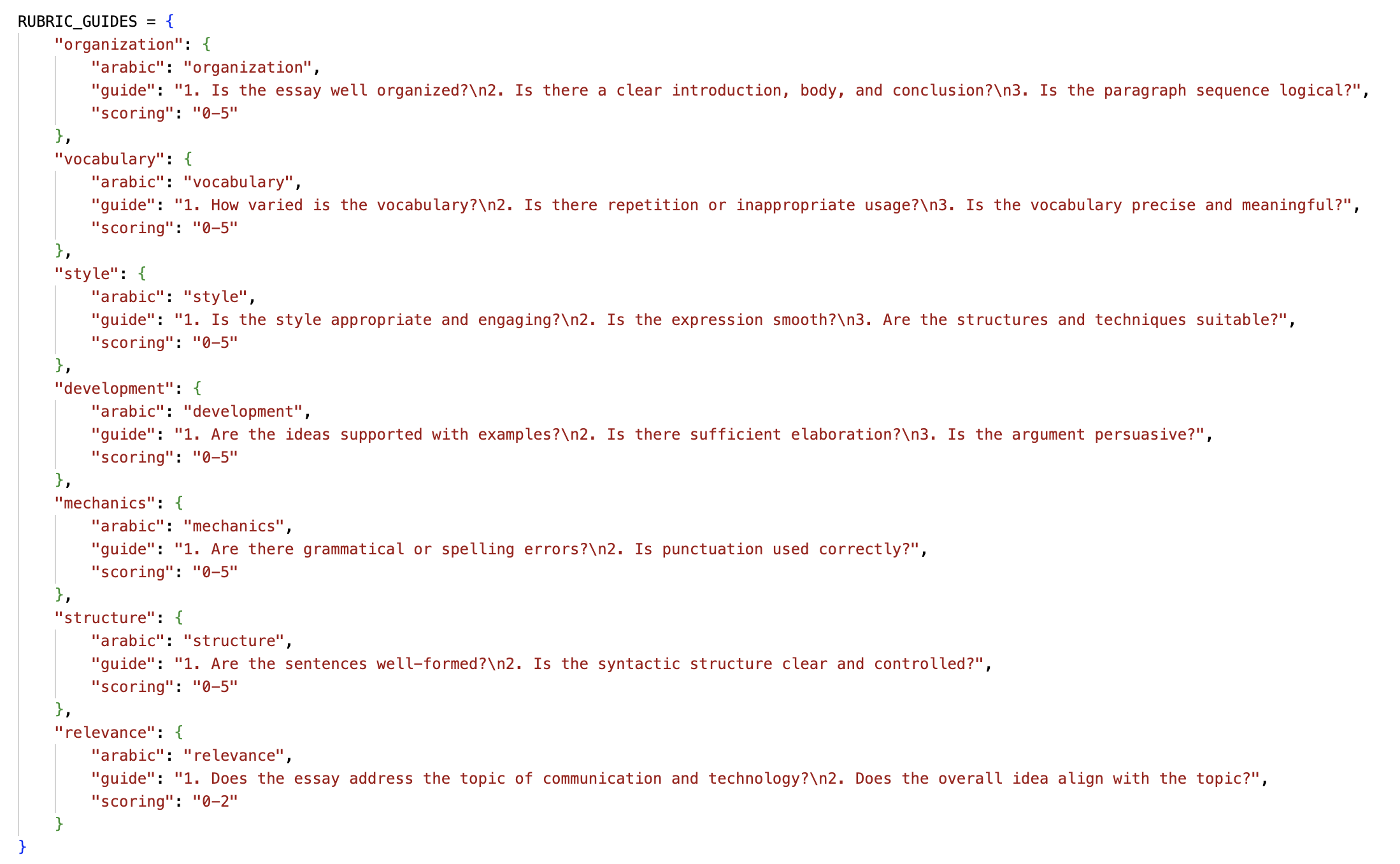}
\caption{Rubric-Guided Few-Shot Prompting - Level 3 (first part)}

\end{figure*}

\begin{figure*}[t]\includegraphics[width=1\linewidth]{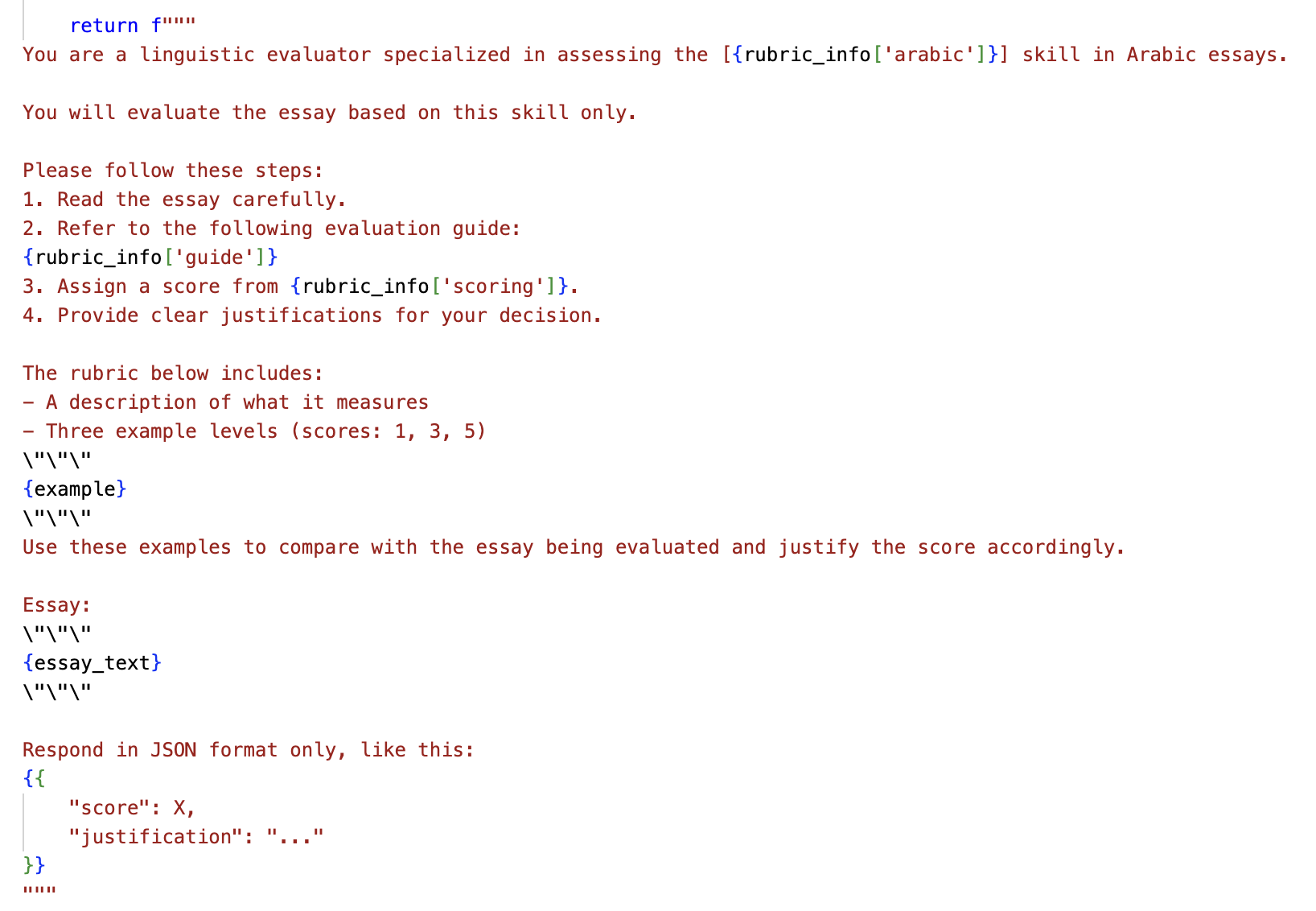}
    \caption{Rubric-Guided Few-Shot Prompting - Level 3 (second part)}
    \hfill

\end{figure*}

\end{document}